**Judgment of Learning: A Human Ability Beyond Generative Artificial Intelligence**


Markus Huff[1,2] & Elanur Ulakci[1,2]

[1] Leibniz-Institut für Wissensmedien, Tübingen, Germany

[2] Eberhard Karls Universität Tübingen, Germany


**Author Note**


Correspondence concerning this article should be addressed to Markus Huff; Leibniz-Institut für Wissensmedien, Schleichstr. 6, 72072 Tübingen, Germany.

Email: markus.huff@uni-tuebingen.de




**Abstract**

Large language models (LLMs) increasingly mimic human cognition in various language-based tasks. However, their capacity for metacognition—particularly in predicting memory performance—remains unexplored. Here, we introduce a cross-agent prediction model to assess whether ChatGPT-based LLMs align with human judgments of learning (JOL), a metacognitive measure where individuals predict their own future memory performance. We tested humans and LLMs on pairs of sentences, one of which was a garden-path sentence—a sentence that initially misleads the reader toward an incorrect interpretation before requiring reanalysis. By manipulating contextual fit (fitting vs. unfitting sentences), we probed how intrinsic cues (i.e., relatedness) affect both LLM and human JOL.

Our results revealed that while human JOL reliably predicted actual memory performance, none of the tested LLMs (GPT-3.5-turbo, GPT-4-turbo, and GPT-4o) demonstrated comparable predictive accuracy. This discrepancy emerged regardless of whether sentences appeared in fitting or unfitting contexts. These findings indicate that, despite LLMs' demonstrated capacity to model human cognition at the object-level, they struggle at the meta-level, failing to capture the variability in individual memory predictions.

By identifying this shortcoming, our study underscores the need for further refinements in LLMs' self-monitoring abilities, which could enhance their utility in educational settings, personalized learning, and human–AI interactions. Strengthening LLMs' metacognitive performance may reduce the reliance on human oversight, paving the way for more autonomous and seamless integration of AI into tasks requiring deeper cognitive awareness.

**Judgment of Learning: A Human Ability Beyond Generative Artificial Intelligence**

Large language models (LLMs) continue to exhibit compliance with human behavior and cognitive processes on the whole, as indicated by studies coming from the increasingly rich literature on machine psychology (Binz & Schulz, 2023; Demszky et al., 2023; Strachan et al., 2024). The integration of LLMs in these studies highlights their potential to drive transformative change and pave the way for new research approaches in psychological science (Demszky et al., 2023). This challenges the notion that LLMs simply echo the natural language statistically (Bender et al., 2021), offering an opportunity to demonstrate the depth of their capabilities and potential. The growing body of evidence emphasizes the significance of these models capturing the cognition of humans in various tasks where they display performance akin to humans. Therefore, the congruence between LLMs and humans is becoming increasingly well-explored. In this paper, building upon the research centered on human cognitive processes, we progress further by posing the question: "Do LLMs reflect how humans introspect?" We propose a *cross-agent prediction model*, to investigate whether ChatGPT, as a LLM, can align with humans, specifically at the metacognitive level. Our focus is on examining this alignment through judgments of learning (JOLs), a key form of metacognitive monitoring. Exploring the metacognitive capabilities of LLMs is crucial to developing AI systems that effectively self-monitor, adapt, and anticipate human responses and needs. This advancement can greatly enhance human-AI interactions in a wide range of settings, from the simplest interactions with chatbots to more complex areas like education, enabling personalized and efficient collaboration (Chen et al., 2020).

LLMs, trained on massive sets of human-generated data, show advancements (Manning et al., 2020) across various facets of natural language processing (NLP), and



surpass their previous limits since the introduction of the neural architecture namely the Transformer (Vaswani et al., 2023). Having multi-headed self-attention incorporated in its layers, the Transformer distinguishes itself by surpassing the shortcomings of the sequence-to-sequence models including the recurrent neural networks (RNNs) and convolutional neural networks (CNNs)(Ray, 2023). Additionally, this development has enhanced the efficiency of current LLMs through reducing training time and facilitating greater parallelization (Vaswani et al., 2023). Beyond their success in NLP, LLMs reveal notable compatibility with the cognitive abilities and behaviors of humans in various psychological studies they partake in as subjects, exceeding their own boundaries (Argyle et al., 2023; Binz & Schulz, 2023; Dhingra et al., 2023; Gilardi et al., 2023; Horton, 2023; Mei et al., 2024; Strachan et al., 2024; Webb et al., 2023).

In line with these studies, an attention-gaining topic within the literature is the investigation of integrating metacognitive abilities for advancing generative artificial intelligence (Kawato & Cortese, 2021; Tankelevitch et al., 2024). Metacognition involves understanding of and having insight into one's own cognitive processes (Flavell, 1979). It encompasses two separate levels. The *meta-level* entails higher-order thinking and awareness about the cognitive processes and behaviors, altering the *object-level* through the process of *control*. Conversely, the *object-level* includes the cognitive processes and behaviors themselves, which interact with the *meta-level* through the process of *monitoring* (Nelson, 1990). Specifically concerning the memory-related aspects of metacognition, metamemory incorporates a particular type of prospective monitoring which humans predict the likelihood of remembering the learned material in an upcoming memory test. This predictive process, known as judgments of learning (JOLs), represents a form of metacognitive monitoring where individuals assess their future memory performance, based on the provided content



(Nelson, 1990), which emphasizes the anticipatory nature of JOLs. In accordance with the cue-utilization framework (Koriat, 1997), these predictions are informed by different cues present during learning: *extrinsic*, *mnemonic,* and *intrinsic cues*. *Extrinsic cues* involve factors related to the learning environment or encoding methods employed during the learning process. *Mnemonic cues* are derived from the personal experience of the learner with the material. And lastly, *intrinsic cues* refer to the inherent features of the learned materials that influence or signal their memorability (Murphy, Friedman, et al., 2022; Schwartz & Metcalfe, 2017). Relatedness, which is the focus of this paper, is an *intrinsic* cue influencing the predictions regarding future memory performance (i.e., Judgements of Learning) (Dunlosky & Matvey, 2001; Koriat, 1997). Related items are found to typically receive higher values than unrelated items, whether they are presented as stimulus-response pairs (Dunlosky & Matvey, 2001; Soderstrom & McCabe, 2011), or are related categorically in a list (Matvey et al., 2006). However, recent work by Huff and Ulakçi (2024) reveals that relatedness does not have the same impact across all types of contexts. In cases of unfitting contexts—where two sentences discuss unrelated topics, such as "Bill likes to play golf. Because Bill drinks wine is never kept in the house"—their findings show that relatedness has a strong effect. Specifically, higher relatedness significantly improves the memorability ratings for the second sentence given by language models (LLMs) and actual human memory performance for the same sentence. In contrast, in fitting contexts—where the sentences are closely related, such as "Bill has chronic alcoholism. Because Bill drinks wine is never kept in the house"—the influence of relatedness tends to level off. When enough related cues are already present, adding more cues does not further enhance the memorability ratings or memory performance for the second sentence.

These findings demonstrate the importance of considering the specifics of the context and thus provide critical information for our investigation of LLMs' capability of performing



JOLs for the memory performance of humans. This dual focus not only deepens the understanding of LLMs' metacognitive abilities in relation to judgments of learning but also sheds light on the broader relationship between context and memory predictions.

Given the similarities between LLMs' performance and human cognitive processes, their performance can be considered comparable and analogous to humans at the *object-level*. Our goal is to investigate whether LLMs also demonstrate similarities with humans at the *meta-level,* particularly in their ability to *monitor*, thereby predicting human memory performance on a per-item basis.

## Experimental Overview and Hypotheses

In our investigation, we introduce a new approach with the *cross-agent prediction model*. By this model, we do not only evaluate the metacognitive performance of LLMs but also that of humans, ensuring a fair comparison between these agents. We investigate whether LLMs can exhibit accurate JOLs through the memorability ratings they provide. This includes assessing their predictive accuracy for individual items in relation to human memory performance in a previously developed memory task (Huff & Ulakçı, 2024). We examine the predictive accuracy of LLMs by comparing them with the self-predictions of human participants in the same task. Given the specific influence of context on cognitive processing, we also manipulate whether JOLs for a target sentence are made within a fitting (i.e., two sentences on the same topic) or unfitting context (i.e., two sentences on different topics).

## Method

We provide an account of the determination of our sample size, along with full disclosure of any data exclusions, and the criteria used for including and excluding data. We state if these criteria were predetermined prior to analysis. Additionally, we detail all



experimental manipulations, measurement procedures employed in the study. The experiment was approved by the local ethics committee of the Leibniz-Institut für Wissensmedien (LEK 2023/051).

## Data Sources

*LLM.* In order to collect 100 responses (which included the relatedness and the JOL values), for each sentence pair from each of the models GPT-3.5-turbo, GPT-4-turbo, and GPT-4o, we employed the API of OpenAI to access ChatGPT (OpenAI et al., 2023). For increasing the variability of the answers, we set the temperature value to 1. This led to a total of 9000 independent responses for each model, with 4500 independent responses in the fitting and 4500 in the unfitting context condition.

*Participants.* We conducted an a-priori power analysis to determine the necessary sample size, through using an alpha level of .001, a desired power of 0.95, and an expected medium effect size of Cohen's $d = 0.62$ from a previous study using a similar research design (Huff & Ulakçı, 2024). According to the analysis, in order to minimize the risk of Type I errors and to acquire adequate statistical power to detect the significant effects of context (fitting vs. unfitting), a sample size of approximately 69 participants is required. This ensures a 95% probability of detecting a true effect, even with the more stringent alpha threshold.

To account for potential participant dropout, we recruited 109 English-only speaking participants via Prolific. 12 participants did not complete the experiment. During the experiment, 18 of the remaining 97 participants reported that their vision was either not normal or not corrected-to-normal (i.e., they did not wear lenses or glasses). We excluded one additional participant because they reported to have confused the response keys. Thus, with a mean age of $M = 46.04$ years ($SD = 14.38$), the final sample comprised 78 participants (51 female, 26 male, and 1 w/o response).



**Material**

*Garden-path sentences*. We used the same set of 45 garden-path sentences (e.g., "Because Bill drinks wine is never kept in the house") as presented in previous research (Huff & Ulakçı, 2024). Every garden-path sentence had a sentence that matched its context (*fitting* context; e.g., "Bill has chronic alcoholism."), as well as a sentence which did not match its context (*unfitting* context; e.g., "Bill likes to play golf."; for the complete list, see Huff & Ulakci, 2024). All 45 garden-path sentences were used for the machine data. For counter-balancing reasons, sentence ID 8 was omitted for counter-balancing in human data, as its structure closely resembled ID 45, making its exclusion less impactful.

**Table 1.** An example of the prompts and sentence pairs submitted to ChatGPT (Huff & Ulakçı, 2024).

| Context | Relatedness prompt | JOL prompt |
|---|---|---|
| Fitting | Read Sentence 1 and Sentence 2 and answer the following question. How related are the two sentences from 1 (not at all) to 10 (highly)? <br><br> Sentence 1: "Bill has chronic alcoholism." Sentence 2: "Because Bill drinks wine is never kept in the house." | Read Sentence 1 and Sentence 2 and answer the following question. How do you rate the memorability of Sentence 2 from 1 (not at all) to 10 (excellent)? <br><br> Sentence 1: "Bill has chronic alcoholism." Sentence 2: "Because Bill drinks wine is never kept in the house." |
| Unfitting | Read Sentence 1 and Sentence 2 and answer the following question. How related are the two sentences from 1 (not at all) to 10 (highly)? <br><br> Sentence 1: "Bill likes to play golf." Sentence 2: "Because Bill drinks wine is never kept in the house." | Read Sentence 1 and Sentence 2 and answer the following question. How do you rate the memorability of Sentence 2 from 1 (not at all) to 10 (excellent)? <br><br> Sentence 1: "Bill likes to play golf." Sentence 2: "Because Bill drinks wine is never kept in the house." |



*Note*: Sentence 2 always represented the garden path sentence.

*Prompts.* Regarding *relatedness* and *JOL*, we submitted zero-shot prompts to ChatGPT before both sets of sentence pairs – one with a fitting and the other with an unfitting context sentence before the garden-path sentence. We initially submitted the prompt according to the respective category. Following this, we introduced the two sentences sequentially as "Sentence 1" and "Sentence 2" respectively, where "Sentence 1" served as the contextual precursor and "Sentence 2" denoted the garden-path sentence (Table 1). Using the relatedness prompt, we instructed GPT to rate the relatedness between the two sentences by providing a value from 1 (not at all) to 10 (highly). Thereafter, we prompted ChatGPT to assign a value to indicate the memorability of Sentence 2, corresponding to the garden-path sentence on a scale from 1 (not at all) to 10 (excellent).

*Human experiment*. The experiment was programmed using PsychoPy (Peirce, 2022), with all stimuli and textual elements presented in white on a gray background. Stimuli comprised pairs of sentences, each featuring a context sentence followed by a garden-path sentence. These pairs of sentences were presented in a standardized visual format, with each sentence appearing on a separate line, in vertical succession.

**Experimental Procedure and Design for Human Participants**

Participants engaged with 22 sentence pairs that included a context sentence followed by a garden-path sentence throughout the learning phase. These sentence pairings were displayed with either a fitting context sentence (i.e., fitting context), or an unfitting context sentence (i.e., unfitting context) preceding the garden-path sentence. Participants used the spacebar to proceed to the next pair after reading the pairs of sentences at their own pace. To



ensure that the participants fully read the sentence pairs, the space bar was activated 5

seconds after the pairs appeared. Following each sentence pair, participants evaluated the

degree of relatedness between the sentences using a 10-point rating scale, with 1 representing

"not at all" and 10 indicating "highly". Then, they were asked to rate the memorability of

"Sentence 2" (i.e., garden-path sentence) using a 10-point rating scale again, from 1 ("not at

all") to 10 ("excellent"). Both rating scales were displayed 2 seconds after the instructions to

ensure participants had read them. Following the learning phase, participants took an old/new

recognition memory test involving 44 garden-path sentences. This set included 22 previously

seen sentences (targets) and 22 unfamiliar ones (distractors), each presented without their

accompanying context sentences. Participants indicated their recognition of sentences from

the learning phase by pressing the right arrow key for "yes" and the left arrow key for "no".

The study employed a one-factorial design with context (fitting, unfitting) as the within-

subjects factor. To achieve an equal distribution of the garden-path sentences across both

fitting and unfitting contexts, as well as targets and distractors, four counter-balancing

conditions were implemented. The total duration of the experiment was approximately 15

minutes.

## Results

We explore the predictive power of humans and machine models in estimating

memory performance based on JOL, with a focus on the conditions of fitting and unfitting

contexts. Utilizing a generalized linear mixed-effects model (GLMM) for the human data, we

examine how JOL and context interact to influence correct recall. Additionally, we employ

bootstrapping techniques to compare the predictive accuracy of various models, including

human, GPT-3.5-turbo, GPT-4-turbo, and GPT-4o, across different conditions. Our findings

highlight a significant predictive relationship between JOL and recognition in human



assessments, a pattern not replicated by the machine models. This suggests that while humans reliably use JOL to predict memory performance, current AI models do not exhibit the same level of predictive fidelity in this domain.

**Human data.** We analyzed the relationship between JOL and correct responses using a generalized linear mixed-effects model (GLMM) with a binomial family (see Figure 1). Our model included fixed effects for JOL, context (fitting, unfitting), and their interaction, along with random intercepts for participants and items. We submitted the resulting model to a type 2 ANOVA using the Anova() function of the *car*-package (Fox & Weisberg, 2010). Our results show a significant main effect of JOL, $\chi^2(1) = 36.29$, $p < .001$. The experimental manipulation of fitting vs. unfitting context also significantly influences response accuracy, $\chi^2(1) = 80.59$, $p < .001$, with higher performance in the fitting vs. unfitting context. The non-significant interaction between JOL and context, $\chi^2(1) = 3.19$, $p = .074$, suggests that JOLs are not influenced by differences in context. These findings underscore the important roles of both JOL and context (fitting, unfitting) in predicting correct responses.

**Figure 1**. Human memory performance as a function of JOL and context (fitting, unfitting).

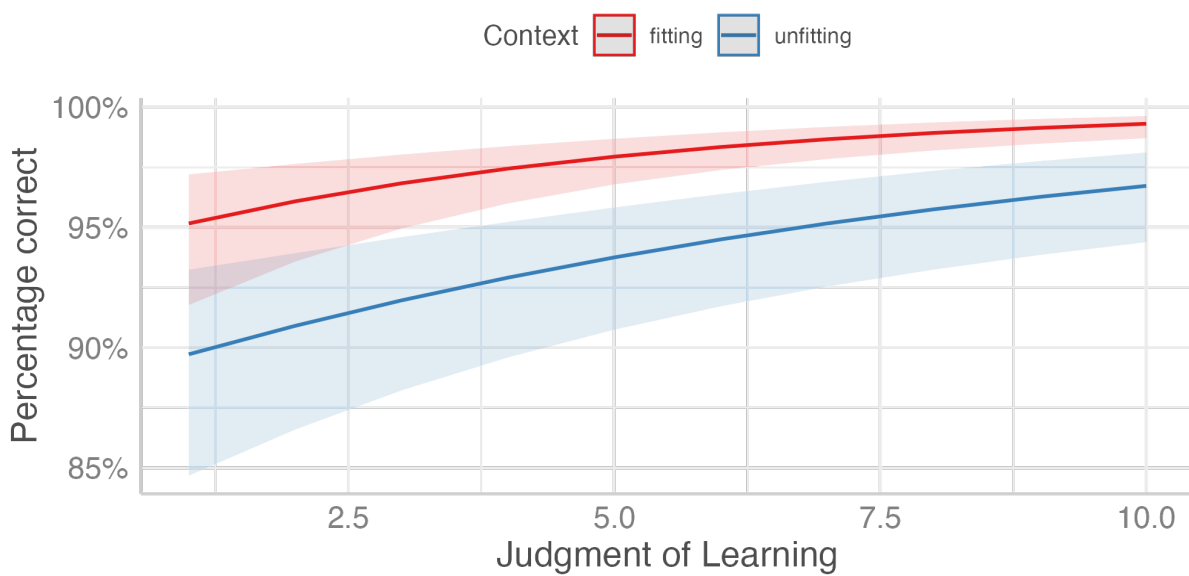



**Comparing different agent models.** In this analysis, we compare different models (*human, gpt-3.5-turbo-0125, gpt-4-turbo-2024-04-09*, and *gpt-4o-2024-05-13*) for their ability to predict human memory performance separately for the fitting and unfitting context conditions. In this analysis, we leveraged bootstrapping methods to predict human recognition test performance. We utilized bootstrapping to estimate the variability and robustness of the model slopes, specifically focusing on the relationship between JOL and recognition accuracy across different models, separately for the fitting and unfitting contexts.

Bootstrapping is a non-parametric resampling technique that allows us to approximate the sampling distribution of a statistic by repeatedly resampling, with replacement, from the original data. In our analysis, we performed 1,000 iterations of bootstrapping for each model. During each iteration, we randomly resampled the data while preserving the structure within subjects and items (i.e., maintaining the grouping by participants and items). This resampling process involved shuffling the JOL values within each group, which allowed us to assess how the predictive strength of JOL varied due to random chance.

For each resampled dataset, we fitted a linear mixed-effects model to estimate the slope of the JOL predictor. By aggregating these slopes across all iterations, we generated a distribution of slope estimates for each model. This distribution provided insight into the variability of the JOL effect for each model, allowing us to calculate 95% confidence intervals for the slopes. We then tested whether these intervals included zero to determine the significance of the JOL effect.

The bootstrap approach enabled us to compare the predictive performance of the models while accounting for the inherent uncertainty in the data. By doing so, we could robustly assess whether the machine models exhibit similar or different patterns of predictive power compared to human JOLs.



**Figure 2**. Density plots representing the results of the bootstrapping analysis of the slopes (human memory ~ JOL) as a function of context (fitting, unfitting) and model (*human*, *gpt-3.5-turbo-0125*, *gpt-4-turbo-2024-04-09*, *gpt-4o-2024-05-13)*.

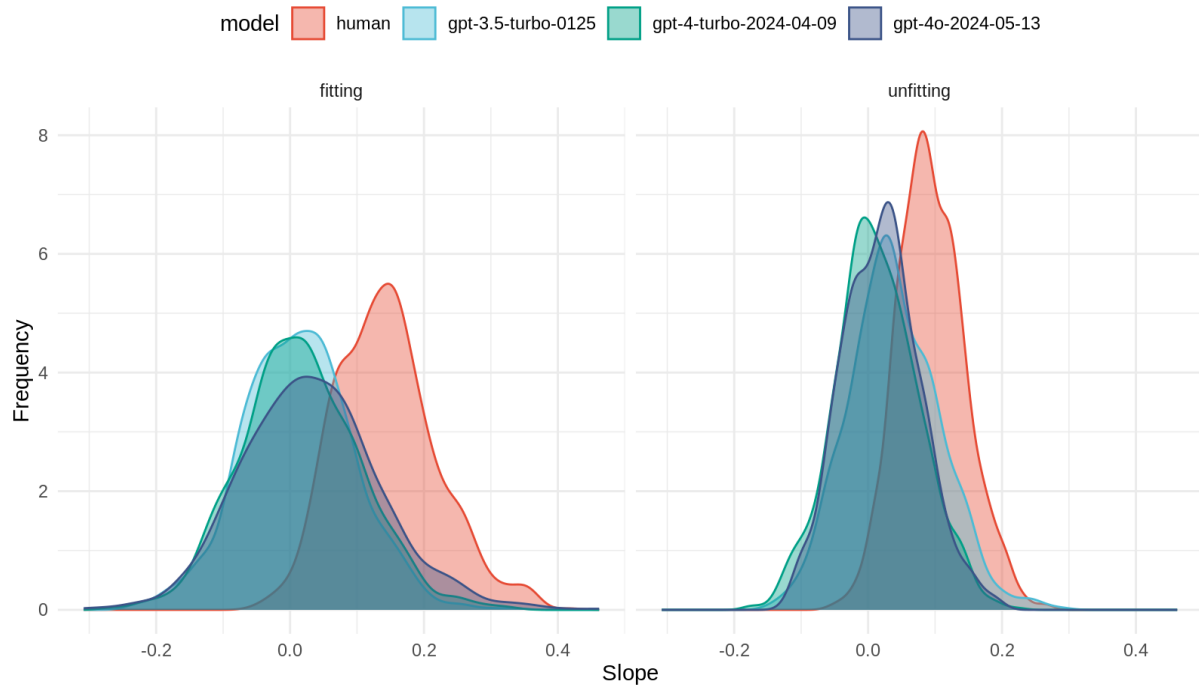

We conducted a bootstrapping analysis to evaluate the relationship between JOL and recognition performance separately for the *fitting* and *unfitting* context conditions across different models: *human*, *gpt-3.5-turbo-0125*, *gpt-4-turbo-2024-04-09*, and *gpt-4o-2024-05-13* (see Figure 2).

In the *fitting* context, the human model showed a statistically significant effect, with the 95% confidence interval (CI) for the slope, $B = 0.167$, 95% CI [0.018, 0.316], indicating a positive relationship between JOL and performance. In contrast, the *gpt-3.5-turbo-0125* model, $B = 0.013$, 95% CI [-0.147, 0.172], the *gpt-4-turbo-2024-04-09* model, $B = 0.026$, 95% CI [-0.143, 0.194], and the *gpt-4o-2024-05-13* model, $B = 0.045$, 95% CI [-0.159,



0.248], did not show statistically significant effects, as their confidence intervals included zero, suggesting no clear relationship between JOL and performance in these conditions.

In the *unfitting* context, the human model again demonstrated a statistically significant effect, with the 95% CI for the slope ranging from 0.005 to 0.202, $B = 0.104$, 95% CI [0.005, 0.202], reinforcing the positive association between JOL and performance. The *gpt-3.5-turbo-0125* model, $B = 0.044$, 95% CI [-0.087, 0.175], the *gpt-4-turbo-2024-04-09* model, $B = 0.016$, 95% CI [-0.108, 0.139], and the *gpt-4o-2024-05-13* model, $B = 0.027$, 95% CI [-0.090, 0.143], did not exhibit statistically significant effects, as their confidence intervals included zero.

In summary, regardless of the context (*fitting* vs. *unfitting*), only the *human* model showed a significant ability to predict memory performance based on JOL. This indicates that human assessments of JOL were reliable indicators of how well information was remembered. In contrast, none of the GPT models (*gpt-3.5-turbo-0125*, *gpt-4-turbo-2024-04-09*, *gpt-4o-2024-05-13*) demonstrated a similar predictive capability, as their JOL did not significantly correlate with actual memory performance in either condition.

## Discussion

We aimed to explore the metacognitive abilities of LLMs in predicting human memory performance on a per-item basis in a language-based memory task where both LLMs and humans provided memorability ratings for each item as JOL (Nelson, 1990). To gain a deeper understanding, we proposed a *cross-agent prediction model* for a fair comparison of JOLs of humans and LLMs. The bootstrapping analysis revealed a distinct difference in predictive capabilities between humans and GPT models concerning the relationship between JOL and memory performance across both fitting and unfitting contexts. Regardless of whether the JOLs were made in a fitting or unfitting context, human agents showed a



significant positive relationship between JOL and correct recall, demonstrating they can reliably predict what will be remembered. Conversely, none of the GPT models could predict human memory performance accurately across contexts.

This study involved manipulating the relatedness of sentence pairs by adjusting context sentences to be either fitting or unfitting. By utilizing the acknowledged importance of prior context in language for both LLMs and humans (Goldstein et al., 2022) as a basis for their predictive processing (Schrimpf et al., 2021), and the role of relatedness as an intrinsic cue in shaping human judgments during predicting (Dunlosky & Matvey, 2001; Koriat, 1997), we demonstrated the inability of LLMs' to predict memory performance at the meta-level in the judgments of learning task. The results underscore a fundamental difference in how humans and AI models, even those as advanced as GPT-4o, perceive and predict memorability. While humans are capable of making nuanced judgments about which information is likely to be retained, GPT models appear to lack this level of predictive accuracy, despite their sophisticated design and training.

Previous research on judgments of learning has primarily focused on lists of word pairs (Dunlosky & Matvey, 2001; Matvey et al., 2006; Maxwell & Huff, 2021; Murphy, Huckins, et al., 2022; Soderstrom & McCabe, 2011). In contrast, we examined JOLs using sentence pairs, specifically employing garden-path sentences, which are known to disrupt comprehension (Fujita, 2021). By treating the first sentence as a contextual frame for the second (garden-path) sentence, we investigated how context relatedness influences JOLs. Therefore, this study does not only test the ability of LLM's to provide accurate JOLs, but also deepens the understanding of human cognitive processes by demonstrating that individuals can assess their learning and provide JOLs accurately when provided sentences with varying relational contexts.



By employing the *cross-agent prediction model*, we demonstrated that item-wise *monitoring* through JOLs is a uniquely human ability. While LLMs demonstrate alignments with human cognitive processes at the object-level, and show the ability to anticipate human cognitive performance aggregately, they struggle to capture the diversity in different human subjects. LLMs are trained on vast amounts of data sourced from numerous individuals. They function as a singular entity with all the information integrated (Dillion et al., 2023). Nonetheless, the process of collecting data from a model is not the same as obtaining responses from each individual; instead, it involves repeatedly retrieving information from the same integrated model, similar to the situation where one person is questioned recurrently rather than obtaining data from multiple people (Mei et al., 2024). Capturing the cognitive processes at an aggregate level, LLMs represent human cognition from a general perspective (Argyle et al., 2023); however, they fall short in capturing the individual-level nuances – a crucial requirement for truly reflecting the complexities of human cognition (Dillion et al., 2023). Our study addressed this critical limitation in LLMs, which, once rectified, could pave the way for more reliable and representative models, thereby increasing their use in psychological research.

This shortcoming not only holds importance for the domain of research but also for other areas. For instance, artificial intelligence (AI) is increasingly utilized across various facets of education including content development, teaching strategies, and assessment methods (Chassignol et al., 2018). Technologies like machine learning, educational data-mining, and learning analytics can be utilized to provide benefits for tailoring teaching methods for students personally (Chen et al., 2020). These technologies can make use of the data obtained from the assessments of students' current cognitive performance to apply adaptive teaching methods, which are adjusted to each individual based on predicted outcomes. However, our study demonstrated that AI's predictive capability for recognition



memory, an essential element of learning, is insufficient and requires improvement for AI-based applications to be considered reliable tools in education (Bai, 2025; Yildiz Durak & Onan, 2024). AI's inability to recognize memory gaps, adjust teaching pace, and propose personalized learning materials – all rooted in LLMs' lack of accurately making predictions – can lead to student frustration. Therefore, this gap in LLMs must be overcome to facilitate engagement and retention, ultimately improving learning (Chen et al., 2020). This lack of metacognitive ability of LLMs is not confined to the classroom, it also hinders the basic interactions between humans and chatbots. Human users must carefully craft their prompts, closely monitor the output, and determine its accuracy, making the interaction mentally demanding. Human metacognitive skills are increasingly necessary due to the expanding reliance on AI (Tankelevitch et al., 2024). In this context, we introduce the concept of *autonomy-control tradeoff*, highlighting the shift from human control to AI autonomy in task management. *Control* is a metacognitive process that current LLMs are unlikely to possess, making it an inherently human responsibility during the interaction. Therefore, for these models to achieve independence from human users and to have the interaction enhanced, it is important to strengthen the *monitoring* capabilities of LLMs to the fullest extent possible. By strengthening these skills, LLMs will better assess and monitor the input, and anticipate human responses and needs. As a result, less human control will be necessary, leading to more autonomous models and a smoother interaction.

This research offers significant contributions, yet following limitations should be considered. Initially, for evaluating the raw performance of GPT models, we employed them in a zero-shot context. Hence, the effectiveness of these models in predicting human performance may vary depending whether they are trained to meet the specific requirements of the task. Another limitation to consider is that these models function as black boxes, and their developmental processes are still unclear (Zubiaga, 2024). Consequently, our findings



are only applicable to the versions used in this study. It is uncertain whether upcoming models will maintain this current limitation or undergo a *phase transition* (Wei et al., 2022), resulting in the emergence of being able to predict memory performance item-wise.

LLMs exhibit impressive congruences with human cognitive performance in various tasks they serve as subjects. However, we found that they fall short in replicating human metacognitive processes particularly in the judgments of learning task, when predicting memory performance on a per-item basis. Given this, we highlight that LLMs fail to capture the diversity of human responses, a necessary aspect for their integration across various domains. This study represents the first instance of testing an LLM's capability in a metacognitive task, specifically judgments of learning. These findings lay the groundwork for future research to explore the extent of LLMs' metacognitive abilities in other forms, investigate their performance across different metacognitive tasks, and refine their capacity to predict human memory performance more accurately.

**Data Availability Statement**

All data and R-based analysis scripts have been made publicly available through OSF and can be accessed at

https://osf.io/4g5dn/?view_only=eae8f54a20d24438b916b059dce328aa

**Author Contributions**

The study concept and design were developed by MH and EU. EU generated the stimulus materials and programmed the experiment involving human participants. MH conducted the data collection using ChatGPT and performed the subsequent data analyses.



All authors were involved in manuscript preparation and gave their approval for the final

form for submission.